\documentclass[letterpaper]{article} 
\usepackage{aaai2026}  
\usepackage{times}  
\usepackage{helvet}  
\usepackage{courier}  
\usepackage[hyphens]{url}  
\usepackage{graphicx} 
\def\UrlFont{\rm}  
\usepackage{natbib}  
\usepackage{caption} 
\usepackage{amsmath,amssymb}
\usepackage{algorithm}
\usepackage{algpseudocode}
\usepackage{float} 
\usepackage{dsfont}
\usepackage{booktabs}
\usepackage{threeparttable}
\usepackage{graphicx}   
\usepackage{subcaption} 
\usepackage{tabularx}

\usepackage{newfloat}
\usepackage{listings}
\DeclareCaptionStyle{ruled}{labelfont=normalfont,labelsep=colon,strut=off} 
\floatstyle{ruled}
\newfloat{listing}{tb}{lst}{}
\floatname{listing}{Listing}
\title{Filtered-ViT: A Robust Defense Against Multiple Adversarial Patch Attacks}
\author{
    Aja Khanal,
    Ahmed Faid,
    Apurva Narayan
}
\affiliations{
    Department of Computer Science, University of Western Ontario\\

    1151 Richmond St,\\
    London, ON N6A 3K7, Canada\\
    akhanal3@uwo.ca, afaid@uwo.ca, apurva.narayan@uwo.ca
}

\usepackage{bibentry}

\begin{document}

\maketitle

\begin{abstract}
Deep learning vision systems are increasingly deployed in safety-critical domains such as healthcare, yet they remain vulnerable to small adversarial patches that can trigger misclassifications. Most existing defenses assume a single patch and fail when multiple localized disruptions occur, the type of scenario adversaries and real-world artifacts often exploit. We propose Filtered-ViT, a new vision transformer architecture that integrates SMART Vector Median Filtering (SMART-VMF), a spatially adaptive, multi-scale, robustness-aware mechanism that enables selective suppression of corrupted regions while preserving semantic detail. On ImageNet with LaVAN multi-patch attacks, Filtered-ViT achieves 79.8\% clean accuracy and 46.3\% robust accuracy under four simultaneous 1\% patches, outperforming existing defenses. Beyond synthetic benchmarks, a real-world case study on radiographic medical imagery shows that Filtered-ViT mitigates natural artifacts such as occlusions and scanner noise without degrading diagnostic content. This establishes Filtered-ViT as the first transformer to demonstrate unified robustness against both adversarial and naturally occurring patch-like disruptions, charting a path toward reliable vision systems in truly high-stakes environments.
\end{abstract}


\section{Introduction}
Machine learning models are increasingly deployed in safety-critical settings such as autonomous driving, surveillance, and importantly, medical imaging. Despite their success on standard benchmarks, these models remain alarmingly fragile. A small localized perturbation can cause a classifier to fail, creating risks that range from misidentifying a stop sign in traffic~\cite{sitawarin2018darts, lu2017notfooled} to misclassifying a tumor in a radiograph~\cite{ma2021understanding}. Among the most practical and damaging threats are adversarial patch attacks, where a small region of an image is corrupted to force a misclassification. Unlike imperceptible perturbations~\cite{szegedy2014intriguing, goodfellow2015explaining}, patches are physically realizable~\cite{brown2017adversarial, karmon2018lavan}, require no access to the model, and present a serious barrier to trustworthy deployment.  

Most existing defenses assume a single adversarial patch. In practice, however, adversaries or natural imaging artifacts often introduce multiple corrupted regions simultaneously. Our experiments show that even leading defenses such as Smoothed-ViT~\cite{salman2022certified}, ECViT~\cite{qian2025ecvit}, and ResNet50 with derandomized smoothing~\cite{levine2020derandomized} lose significant accuracy when scaling from one to four adversarial patches. This exposes a critical gap: current defenses cannot guarantee reliability under multi-patch conditions, which are both realistic and high-risk.  

We address this challenge with \textbf{Filtered-ViT}, a new vision transformer architecture that integrates \textbf{SMART Vector Median Filtering (SMART-VMF)} as a core component. SMART-VMF stands for Spatially-adaptive, Multi-scale, Attention-guided, Robustness-aware, and Task-optimized Vector Median Filtering. Unlike conventional preprocessing filters~\cite{smolka2001fast, bae2018fast}, SMART-VMF is embedded directly into the transformer pipeline, where it selectively suppresses adversarial or artifact-heavy regions while preserving semantic detail. In addition, it leverages derandomized smoothing to create robust ablations~\cite{levine2020derandomized} and builds on the Vision Transformer as a strong base classifier~\cite{dosovitskiy2021image}. Importantly, SMART-VMF requires no significant computational overhead, as the filtering mechanism operates in polynomial time and is capable of running on consumer hardware.  

We evaluate Filtered-ViT against adversarial patches generated using the LaVAN attack~\cite{karmon2018lavan} on the ImageNet dataset~\cite{deng2009imagenet}, and further verify its effectiveness in a real-world case study on radiographic imagery. In both domains, Filtered-ViT not only improves robustness to adversarial corruption but also mitigates natural artifacts such as scanner noise and occlusions without erasing diagnostically relevant features. Compared to state-of-the-art baselines, Filtered-ViT achieves substantially higher performance under multi-patch attacks, with early results showing gains of over 15\% in robust accuracy and over 16\% in clean accuracy when tested against four simultaneous 1\% square patches. This dual evaluation demonstrates that Filtered-ViT is both principled and practical, combining adversarial robustness with reliability in safety-critical applications.

\noindent Our contributions are as follows:
\begin{itemize}
    \item We propose Filtered-ViT, a new transformer architecture that integrates SMART Vector Median Filtering (SMART-VMF) as a robustness-aware, task-optimized mechanism for resisting multi patch-level corruption.  
    \item We conduct extensive evaluation on ImageNet with LaVAN adversarial patches, showing consistent improvements in both clean and robust performance compared to state-of-the-art baselines.  
    \item We verify our approach in a real-world case study on medical imagery, demonstrating that Filtered-ViT mitigates both adversarial artifacts and natural imaging noise without degrading diagnostic fidelity, establishing it as a practical step toward trustworthy medical vision systems.  
\end{itemize}

\section{Background and Related Work}
\subsection{Adversarial Patch}
A seminal approach to creating adversarial patch attacks was presented by Brown et al.~\cite{brown2017adversarial}. The adversarial attack involves taking a small patch \(p\), applying a random transformation \(t\) such as rotation or scaling, and placing it at a location \(l\) on an image \(x \in \mathbb{R}^{w \times h \times c}\), where \((w,h)\) is the resolution and \(c\) is the number of channels. This process is formalized as the application function \(A(p,x,l,t)\). Given an image training set \(X\), a distribution over patch transformations \(T\), a distribution of patch locations \(L\), and some target class \(\hat{y}\), the trained patch \(\hat{p}\) is obtained using the Expectation over Transformation (EOT) framework~\cite{athalye2018synthesizing}:

\begin{equation}
\hat{p}=\arg \max _{p} \mathbb{E}_{x \sim X, t \sim T, l \sim L}\left[\log \operatorname{Pr}(\hat{y} \mid A(p, x, l, t))\right]
\end{equation}

After training, patches are evaluated across ImageNet models such as Inception-v3, ResNet-50, Xception, VGG16, and VGG19, where they outperform black-box attacks in forcing misclassifications. The success of such patches underscores a fundamental vulnerability: networks over-prioritize local salient features and can be manipulated by localized noise that dominates natural structures.

\subsection{LaVAN: Localized and Visible Adversarial Noise}
Karmon et al.~\cite{karmon2018lavan} introduced the LaVAN framework, where small noise patches occupying as little as 2\% of the image are inserted at non-salient regions, yet still induce targeted misclassification. Unlike Brown et al., LaVAN patches need not overlap the main object, making them more insidious. Importantly, they remain transferable across arbitrary images and locations, presenting a stronger real-world threat. These findings highlight the fragility of deep vision systems and their heightened vulnerability to localized perturbations.

\subsection{Vision Transformers (ViTs)}
Transformers, long dominant in NLP~\cite{vaswani2017attention}, have been adapted to vision tasks through the Vision Transformer (ViT)~\cite{dosovitskiy2021image}. ViT splits an image \(x \in \mathbb{R}^{w \times h \times c}\) into non-overlapping patches \(x_p \in \mathbb{R}^{N \times (P^2 \cdot c)}\), where \((P, P)\) is the patch resolution and \(N = \tfrac{hw}{P^2}\). These patches are embedded and passed through a transformer encoder to produce compact image-level representations. ViTs achieve state-of-the-art performance on ImageNet~\cite{deng2009imagenet}, offering both accuracy and interpretability advantages. Given their patch-based inductive bias, ViTs also present a natural backbone for exploring defenses against adversarial patch attacks.

\subsection{(De)randomized Smoothing}
Levine \& Feizi~\cite{levine2020derandomized} proposed a certifiable defense against adversarial patch attacks by exploiting their spatial locality. Two mechanisms are introduced: block smoothing $\Delta_{bk}$ and band smoothing $\Delta_{bd}$. For block smoothing, an \(s \times s\) block is retained from the image, giving \(k = s^2\) pixels. For a patch of size \(m \times m\), the probability that a block intersects with the patch is

\begin{equation}
\Delta_{bk}=\frac{(m+s-1)^{2}}{hw}=\frac{(m+\sqrt{k}-1)^{2}}{hw}<4 \frac{\max (hm, k)}{hw}
\end{equation}

For band smoothing, a row or column of width \(s\) is retained, yielding \(k=sh\) pixels and a corresponding overlap probability:

\begin{equation}
\Delta_{bd}=\frac{m+s-1}{w}=\frac{m+\tfrac{k}{h}-1}{w}<2 \frac{\max (hm, k)}{hw}
\end{equation}

While effective, these methods often reduce clean accuracy and introduce inference overhead. More critically, their guarantees rely on assumptions about patch size and placement that do not always hold in practice, limiting their utility in real-world applications such as medical imaging where artifacts may be irregular, overlapping, or distributed across multiple regions.

\subsection{Smoothed-ViT}
Salman et al.~\cite{salman2022certified} combined randomized smoothing with ViTs to create Smoothed-ViT, achieving significantly higher certified accuracy on ImageNet and CIFAR-10 compared to ResNets. ViTs better tolerate ablated images due to their global attention mechanism, enabling faster and more accurate certified defenses. However, Smoothed-ViT is primarily effective only against single adversarial patches. Its performance drops substantially when multiple or distributed patches are introduced, and its certification guarantees do not extend well to these multi-patch conditions. This creates a gap between theoretical robustness and the realities of adversarial attacks or natural imaging artifacts in clinical radiographs.

\subsection{Vector Median Filter (VMF)}
Median filtering is a classic defense against image corruption, and Vector Median Filters (VMFs)~\cite{smolka2001fast,liu2013noise,bae2018fast} extend this principle to multichannel vectors by treating each pixel as a high-dimensional entity. By minimizing the aggregate distance to neighboring vectors, VMFs are highly effective for impulse and non-Gaussian noise:

\begin{equation}
    \hat{x} = \arg \min_{x_i \in W} \sum_{x_j \in W} \|x_i - x_j\|
\end{equation}

where \(W\) is the local window around each pixel. Although VMFs are powerful for suppressing stationary noise, they are fundamentally static and lack the ability to adapt to structured adversarial perturbations. When applied aggressively, they risk blurring diagnostically important details in medical imagery, undermining their utility in safety-critical applications.  

\subsection{Adversarial Attacks in Medical Imaging}
Recent studies have demonstrated that medical imaging models are highly susceptible to adversarial perturbations, raising critical concerns for clinical deployment. Finlayson et al.~\cite{finlayson2019adversarial} showed that imperceptible perturbations can cause diagnostic models to misclassify dermatology and radiology images, while Mirsky et al.~\cite{mirsky2019ct} revealed that adversarial noise injected into CT scans could manipulate cancer detection outcomes. Beyond imperceptible attacks, Xiao et al.~\cite{xiao2022patch} and Ma et al.~\cite{ma2021understanding} highlighted the risks of patch-based adversarial noise, which can be physically realizable and resemble natural artifacts such as occlusions or scanner defects. These findings underscore that adversarial threats are not merely theoretical but directly endanger reliability in high-stakes medical contexts, motivating defenses that both preserve diagnostic fidelity and withstand localized corruption.

\section{Methodology}
\subsection{Patch Training}
The LaVAN attack is particularly suitable for our evaluation because even a patch as small as 1\% of the original image can reliably induce misclassification. Moreover, its design permits the placement of multiple patches in non-salient regions, making it an especially rigorous test for defenses against multi-patch corruption \cite{karmon2018lavan}.  

Following Karmon et al., we adopt a gradient-based optimization approach to train adversarial patches. For an input image $x \in \mathbb{R}^{w \times h \times c}$, a classifier $M$ assigns class probabilities $\vec{y}=p_{M}(y \mid x)$, where $\vec{y}$ encodes the likelihood of each label. The predicted class is $y=\arg \max _{y^{\prime}} p_{M}(y=y^{\prime} \mid x)$. Let $y_{s}$ be the classifier’s prediction for $x$, and define $x^{\prime}$ as an adversarial image misclassified as a target class $y_{t}$. This adversarial sample can be expressed as $x^{\prime}=x+\delta$, where $\delta \in \mathbb{R}^{n}$ represents the perturbation. The optimization objective is to maximize the membership probability $p_{M}\left(y=y_{t} \mid x+\delta\right)$ by learning $\delta$ through stochastic gradient descent.  

Perturbations are localized using a binary mask $q$, which defines the patch region and ensures element-wise replacement:
\begin{equation}
(1-q) \cdot x+q \cdot \delta, \quad q \in\{0,1\}^{n}
\end{equation}

To accelerate convergence, LaVAN further optimizes a loss function based on network activations before the final softmax layer. This formulation maximizes the target class score while simultaneously suppressing the source class score:
\begin{equation}
\begin{split}
    \arg \max _{\delta} \; p_{M}\Big[M\left(y=y_{t} \mid(1-q) \cdot x+q \cdot \delta\right) \\
    -M\left(y=y_{s} \mid(1-q) \cdot x+q \cdot \delta\right)\Big]
  \end{split}
\end{equation}

The training process for a LaVAN patch attack is summarized below:

\begin{algorithm}[H]
\caption{Training procedure for LaVAN patch attack}
\begin{algorithmic}
\Require image $x$, model $p_{M}$, target class $y_{t}$, target probability $s$, mask $q$
\State $y_{s} \gets \arg\max_{y} p_{M}(y \mid x)$
\State $\delta \gets \vec{0}$
\State $x' \gets (1-q)\cdot x + q\cdot \delta$
\State $i \gets 0$
\While{$p_{M}(y=y_{t}\mid x') < s$ \textbf{and} $i < n_t$}
  \State $L_{t} \gets M(y=y_{t}\mid x')$
  \State $L_{\max} \gets M(y=y_{\max}\mid x')$
  \State $\nabla_{t} \gets \frac{\partial L_{t}}{\partial x}$
  \State $\nabla_{\max} \gets \frac{\partial L_{\max}}{\partial x}$
  \State $\delta \gets \delta - \varepsilon \cdot (\nabla_{t} - \nabla_{\max})$
  \State $x' \gets (1-q)\cdot x + q\cdot \delta$
  \State $i \gets i + 1$
\EndWhile
\end{algorithmic}
\end{algorithm}

\subsection{Patch Placement}

Once trained, adversarial patches are applied to test images by replacing original pixel values at specific locations. To evaluate robustness, we vary both patch number and size across seven setups: one to four patches, each ranging from 1\% to 4\% of the image area. Patch sizes are chosen to reflect realistic conditions, since anomalies or artifacts in radiographs, such as calcifications, sensor noise, or cropping effects, typically occupy only a small fraction of the image but can heavily influence interpretation. As Table~\ref{tab1} shows, the patch size is fixed for each placement setup

Patches are placed at the image corners rather than the center. This placement mimics adversarial strategies that exploit non-salient regions to evade detection, while also aligning with medical scenarios where artifacts often appear at image boundaries. Evaluating defenses under these conditions ensures relevance to both adversarial robustness and real-world medical imaging artifacts.

Formally, for an image $x \in \mathbb{R}^{w \times h \times c}$ and a square patch $p \in \mathbb{R}^{s \times s \times c}$, we define a placement function $P(x, p, n)$, where $n \in \{\mathbb{N} \mid n \leq 4\}$ specifies the number of patches. For the case of four patches, the resulting set of corner locations $L$ is given by:


\begin{multline}\label{eq}
L = P(x,p,4)\\
= \{(0,0),\ (0,\,w-s),\ (h-s,\,0),\ (h-s,\,w-s)\}
\end{multline}

\begin{figure}[htbp]
\centerline{\includegraphics[scale=0.50]{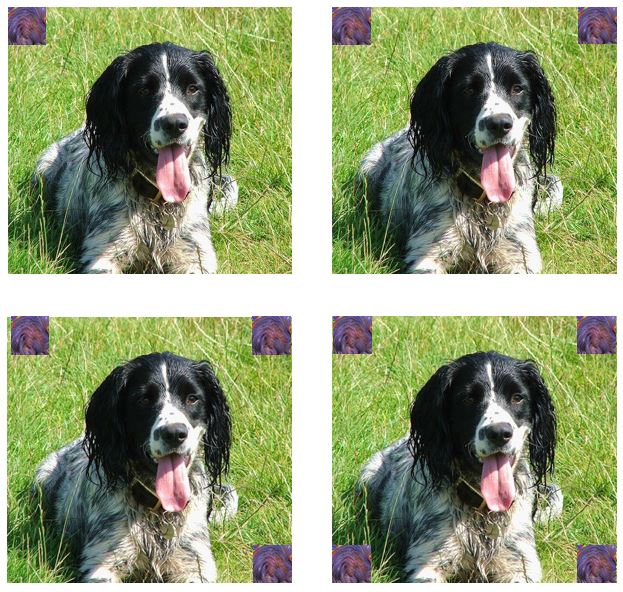}}
\caption{Example of adversarial patches placed at image corners.}
\label{fig}
\end{figure}

An illustration of the multi-patch placement is provided in Figure~\ref{fig}.

\subsection{Filtered-ViT Architecture}\label{AA}

\begin{figure*}
\includegraphics[scale=0.655]{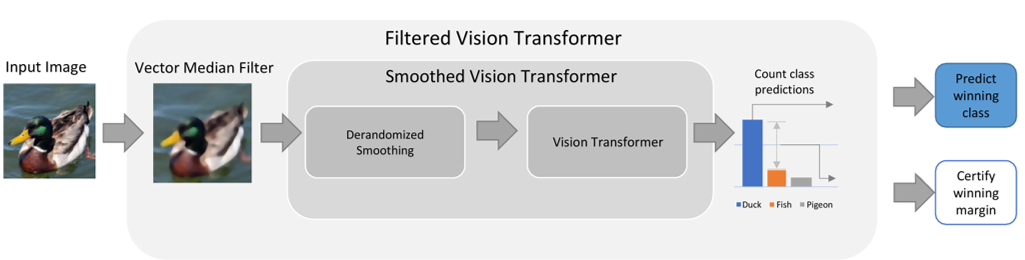}
\caption{Filtered-ViT Architecture: SMART-VMF first filters adversarial or artifact-heavy regions, followed by derandomized smoothing and classification using a Vision Transformer.}
\label{Filtered-ViT Architecture}
\end{figure*}

The architecture of Filtered-ViT integrates adversarial filtering with robust classification in a single pipeline (Figure~\ref{Filtered-ViT Architecture}). Building on Smoothed-ViT \cite{salman2022certified}, Filtered-ViT preserves derandomized smoothing and the Vision Transformer as a strong base classifier, but introduces a critical innovation: the SMART Vector Median Filter (SMART-VMF). 

Standard randomized smoothing provides certified guarantees against $\ell_p$ perturbations by generating randomized ablations of an input image and aggregating predictions. However, it assumes that adversarial noise is diffuse and uniformly distributed, which makes it poorly suited to localized multi-patch attacks. In practice, as we demonstrate, even leading defenses such as Smoothed-ViT and ECViT experience severe accuracy degradation when adversarial patches scale from one to four simultaneous corruptions. This limitation motivates the introduction of a filtering mechanism capable of detecting and suppressing concentrated artifacts before smoothing and classification.

\paragraph{SMART-VMF.}
Unlike static vector median filters that apply uniform filtering across an image, SMART-VMF selectively suppresses adversarial patches while preserving semantically important structures. This is achieved through three mechanisms: (1) adaptive weighting, where filtering strength depends on similarity, spatial proximity, and attention; (2) multi-scale windows, which adapt to different artifact sizes; and (3) residual-driven fusion, which chooses the most reliable scale at each pixel.

Formally, given an input image $x \in \mathbb{R}^{h \times w \times c}$, the SMART-VMF output at pixel $i$ is defined as:
\begin{equation}
\hat{x}_i^{(s)} = \arg\min_{z \in \mathbb{R}^c} \sum_{j \in W_s(i)} w_{ij}\,\|z - x_j\|_2
\end{equation}
where $W_s(i)$ is a window of size $s \times s$ and $w_{ij}$ are adaptive weights:
\begin{equation}
w_{ij} \propto 
\exp\!\left(-\tfrac{\|x_i - x_j\|_2^2}{\sigma_c^2}\right) 
\cdot 
\exp\!\left(-\tfrac{\|p_i - p_j\|_2^2}{\sigma_p^2}\right) 
\cdot (1 + \lambda\,a_j)
\end{equation}
Here $\sigma_c$ and $\sigma_p$ control content and spatial falloff, and $a_j \in [0,1]$ denotes an attention-derived saliency score. We normalize weights so that $\sum_j w_{ij} = 1$.

To capture adversarial artifacts of varying sizes, SMART-VMF computes candidates $\hat{x}_i^{(s)}$ across multiple window scales $s \in \{3,5,7\}$ and assigns each a residual:
\begin{equation}
e_s(i) = \sum_{j \in W_s(i)} w_{ij}\,\|\hat{x}_i^{(s)} - x_j\|_2
\end{equation}
Finally, the outputs are fused by reliability-weighted averaging:
\begin{equation}
v_i = \sum_{s} \pi_s(i)\,\hat{x}_i^{(s)}, 
\quad \pi_s(i) = \frac{\exp(-e_s(i)/\tau)}{\sum_{s'} \exp(-e_{s'}(i)/\tau)}
\end{equation}
where $\tau$ controls the sharpness of the fusion. This ensures that the scale most consistent with local image structure dominates the result.

\begin{algorithm}
\caption{SMART-VMF}
\begin{algorithmic}
\Require image $x \in \mathbb{R}^{h \times w \times c}$, scales $\mathcal{S}=\{3,5,7\}$, attention map $A \in [0,1]^{h \times w}$, variances $\sigma_c,\sigma_p$, attention weight $\lambda$, fusion temperature $\tau$, max iterations $T$, small $\varepsilon$
\Ensure filtered image $v \in \mathbb{R}^{h \times w \times c}$
\For{each pixel $i$ in $x$}
  \For{each $s \in \mathcal{S}$}
    \State $W_s(i) \gets$ square window of side $s$ centered at $i$
    \For{each $j \in W_s(i)$}
      \State $a_j \gets A_j$ if $A$ is given else $a_j \gets 0$
      \State $w_{ij} \gets \exp\!\big(-\|x_i-x_j\|_2^2/\sigma_c^2\big)\cdot \exp\!\big(-\|p_i-p_j\|_2^2/\sigma_p^2\big)\cdot (1+\lambda a_j)$
    \EndFor
    \State Normalize $w_{ij} \gets w_{ij} / \sum_{j \in W_s(i)} w_{ij}$
    \State $z \gets x_i$ \Comment{initialize vector median iterate}
    \For{$t=1$ to $T$}
      \State \textbf{for stability:} $d_j \gets \max\big(\|z-x_j\|_2,\varepsilon\big), \forall j \in W_s(i)$
      \State $z \gets \dfrac{\sum_{j \in W_s(i)} \frac{w_{ij}}{d_j}\,x_j}{\sum_{j \in W_s(i)} \frac{w_{ij}}{d_j}}$ \Comment{Weiszfeld update}
    \EndFor
    \State $\hat{x}_i^{(s)} \gets z$
    \State $e_s(i) \gets \sum_{j \in W_s(i)} w_{ij}\,\|\hat{x}_i^{(s)}-x_j\|_2$ \Comment{residual energy}
  \EndFor
  \State $\pi_s(i) \gets \exp(-e_s(i)/\tau) \big/ \sum_{s' \in \mathcal{S}} \exp(-e_{s'}(i)/\tau)$ \Comment{softmax fusion}
  \State $v_i \gets \sum_{s \in \mathcal{S}} \pi_s(i)\,\hat{x}_i^{(s)}$
\EndFor
\end{algorithmic}
\end{algorithm}

By integrating SMART-VMF directly before derandomized smoothing, Filtered-ViT ensures that adversarial or artifact-heavy regions are attenuated before ablations are sampled. This synergy provides robustness beyond static filtering or smoothing alone. Importantly, while developed on ImageNet with LaVAN patches to establish controlled benchmarks, the architecture scales naturally to medical imagery, where localized artifacts such as scanner noise, occlusions, or adversarial perturbations share a similar structure. This makes Filtered-ViT both theoretically principled and practically applicable in safety-critical domains.

\subsection{Obtaining Robust and Clean Accuracies}

To evaluate Filtered-ViT, we adopt two complementary metrics: \textit{clean accuracy} and \textit{robust accuracy}. These are standard in adversarial machine learning, as they capture both baseline predictive ability and resilience under attack.  

\paragraph{Clean Accuracy.}  
This measures the proportion of correct predictions on unperturbed data, ensuring that filtering and ablation preserve semantic information. For a classifier $f$, input $x$ of class $c$, and ablation set $S_b(x)$ with $n$ smoothed variants $x_a$, clean accuracy $a_c$ is defined as:
\begin{equation}
a_{c}(x, c) = \frac{100\%}{n} \sum_{x_{a} \in S_{b}(x)}^{n} \mathds{1}\{f(x_{a})=c\} \label{eq}
\end{equation}

\paragraph{Robust Accuracy.}  
This evaluates resilience to patch-based attacks, requiring the majority vote over ablations to strongly favor the true class. With $\Delta \in \{\Delta_{\text{block}}, \Delta_{\text{band}}\}$ denoting patch–ablation overlaps, robust accuracy $a_r$ is:
\begin{equation}
a_{r}(x, c) = \frac{100\%}{n} \sum_{x_{a} \in S_{b}(x)}^{n} \mathds{1}\{a_{c}(x_{a}, c) > \max_{c' \neq c}(\cdot) + 2\Delta \} \label{eq}
\end{equation}

\section{Experimental Setup}
\subsection{Hardware}
All experiments were performed on a consumer-grade workstation equipped with an AMD Ryzen 9 7900X 12-core CPU (24 threads, base clock 4.7 GHz), 32 GB RAM, and an NVIDIA GeForce RTX 4080 SUPER GPU with 16 GB VRAM. This setup ensures that our methodology is reproducible without access to specialized high-performance computing infrastructure, reflecting realistic conditions for independent replication.

\subsection{Model Hyperparameters}
To ensure a fair comparison across architectures, all models were trained under a consistent setup unless otherwise noted.

\paragraph{Optimization and Training.} 
We trained all models using AdamW with an initial learning rate of $3 \times 10^{-4}$, weight decay $0.05$, and cosine learning rate scheduling. Formally, at epoch $t$, the learning rate is given by
\begin{equation}
\eta_t = \frac{1}{2}\eta_0\left(1 + \cos\!\left(\frac{t}{T}\pi\right)\right)
\end{equation}
where $\eta_0$ is the base learning rate and $T$ is the total number of epochs. Training was performed for up to 100 epochs with early stopping (patience = 10) based on validation robust accuracy. The batch size was fixed at 128.

\subsection{Datasets}

\paragraph{ImageNet.} 
To benchmark the baseline classification and evaluate robustness against patch-based adversarial attacks, we adopt the ImageNet dataset. ImageNet is a standard large-scale benchmark containing over one million labeled images in a thousand categories, providing both diversity and complexity to evaluate model generalization. The large class space and high intra-class variability make ImageNet an ideal choice for evaluating models trained under adversarial stress. Furthermore, the use of LaVAN-style adversarial patches on ImageNet is motivated by its well-established role as a reference point in adversarial robustness research. Since patch attacks typically exploit localized regions of high saliency, ImageNet’s broad coverage of object-centric imagery allows us to test whether Filtered-ViT can maintain clean and robust accuracy in varied contexts. By grounding our experiments in ImageNet, we ensure comparability with prior work and provide a rigorous testbed for the proposed SMART-VMF filtering.

\paragraph{Medical Images.} 
To demonstrate real-world applicability, we conduct case studies on chest radiographs from the NIH ChestX-ray14 dataset~\cite{wang2017chestxray8} and brain MRIs from the UK Biobank dataset~\cite{alfaro2018image}. Unlike ImageNet, where adversarial patches are artificially injected, medical images naturally contain patch-like artifacts such as pacemaker wires, surgical clips, ink annotations, motion blur, or susceptibility effects. These localized high-contrast disruptions often dominate attention maps in baseline models, leading to misclassifications, such as confusing a pacemaker with a lung mass or susceptibility artifacts for cerebral microbleeds. By analyzing these cases, we highlight how our framework bridges synthetic adversarial robustness with clinically significant artifact resilience.  

\section{Results}
\begin{figure}[htbp]
    \centering
    \includegraphics[width=\columnwidth]{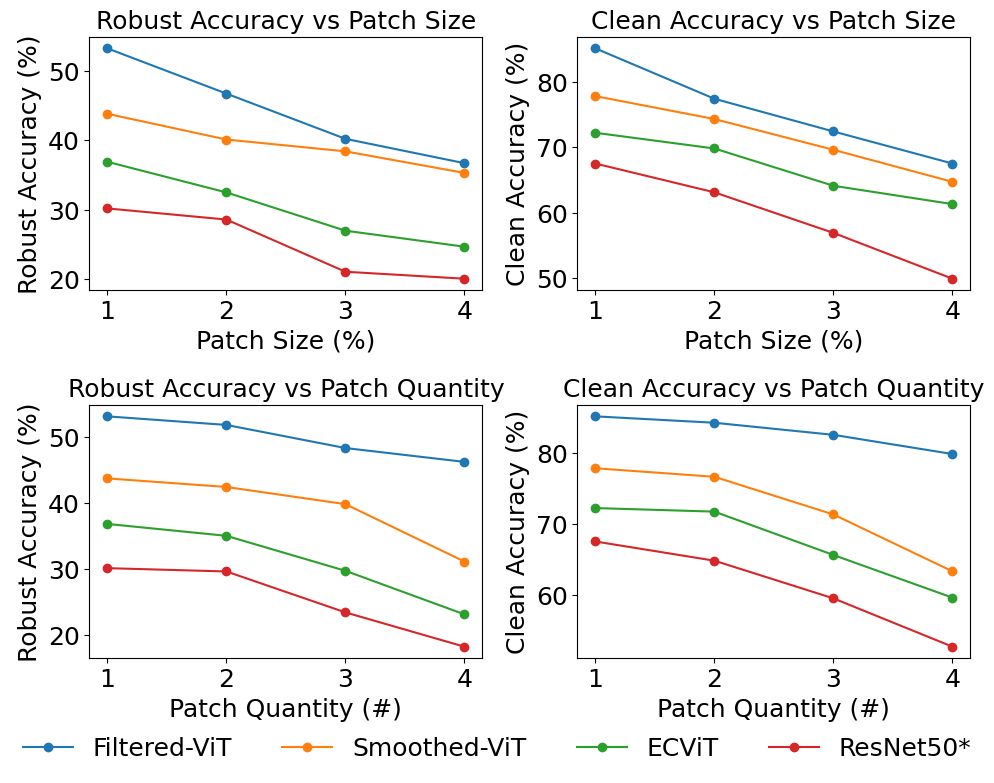}
    \caption{Accuracy trends across patch size and quantity for all models. 
    Top: Robust and Clean accuracy vs. patch size. 
    Bottom: Robust and Clean accuracy vs. patch quantity.}
    \label{fig:patch-results}
\end{figure}

\begin{table}[!t]
\centering
\setlength{\tabcolsep}{3pt} 
\begin{threeparttable}
\begin{tabular}{llcccccccc}
\toprule
 & & \multicolumn{2}{c}{Filtered-ViT} & \multicolumn{2}{c}{Smoothed-ViT} & \multicolumn{2}{c}{ECViT} & \multicolumn{2}{c}{ResNet50*} \\
\cmidrule(lr){3-4} \cmidrule(lr){5-6} \cmidrule(lr){7-8} \cmidrule(lr){9-10}
\# & \% & Rbst & Cln & Rbst & Cln & Rbst & Cln & Rbst & Cln \\
\midrule
1 & $1$ & 53.2 & 85.1 & 43.8 & 77.8 & 36.9 & 72.2 & 30.2 & 67.5 \\
2 & $1$ & 51.9 & 84.2 & 42.5 & 76.6 & 35.1 & 71.7 & 29.7 & 64.8 \\
1 & $2$ & 46.7 & 77.4 & 40.1 & 74.3 & 32.5 & 69.8 & 28.6 & 63.1 \\
3 & $1$ & 48.4 & 82.5 & 39.9 & 71.3 & 29.8 & 65.6 & 23.5 & 59.5 \\
1 & $3$ & 40.2 & 72.4 & 38.4 & 69.6 & 27.0 & 64.1 & 21.1 & 56.9 \\
4 & $1$ & 46.3 & 79.8 & 31.2 & 63.3 & 23.2 & 59.6 & 18.3 & 52.7 \\
1 & $4$ & 36.7 & 67.5 & 35.3 & 64.7 & 24.7 & 61.3 & 20.1 & 49.9 \\
\bottomrule
\end{tabular}
\begin{tablenotes}
\small
\item Note: Despite using a ResNet50 backbone, the derandomized smoothing methodology is still applied to ensure consistency, as all models operate on ablated images.*
\end{tablenotes}
\caption{Clean (Cln) and Robust (Rbst) Accuracies for Various Quantities (\#) and Sizes (\%) on ImageNet ILSVRC 2012 (Validation Set)}
\label{tab1}
\end{threeparttable}
\end{table}
The results presented in Table~\ref{tab1} and Figure~\ref{fig:patch-results} demonstrate that Filtered-ViT provides substantial improvements in both clean and robust accuracy across all patch settings, outperforming Smoothed-ViT, ECViT, and ResNet50*. This advantage becomes particularly evident as the number of adversarial patches increases, which is typically the most difficult scenario for existing architectures. The effectiveness of Filtered-ViT can be directly traced to its design, which integrates selective filtering with the SMART-VMF optimization scheme, addressing specific vulnerabilities that were identified in prior approaches.

\paragraph{Single-Patch Scenarios.}  
For isolated adversarial patches, Filtered-ViT achieves the highest robust and clean accuracies, showing that the model can suppress localized noise without discarding important semantic information. This success arises from the filtering module, which prioritizes high-salience tokens during attention computation. Unlike Smoothed-ViT, which relies on averaging mechanisms that blur signal and noise together, the filtering operation enforces selectivity by actively suppressing tokens with low consistency across feature projections. This is further strengthened by the SMART-VMF objective, which minimizes variance in token representations while maintaining alignment with class-discriminative directions. As a result, perturbations introduced by small patches fail to propagate throughout the attention layers and instead remain localized.

\paragraph{Multi-Patch Scenarios.}  
The superiority of Filtered-ViT becomes most apparent when adversarial patches are distributed across the image. Smoothed-ViT exhibits steep accuracy drops in these conditions because its randomized smoothing does not scale well with multiple localized attacks. Each patch introduces independent noise regions that overwhelm the averaging mechanism. ECViT, which emphasizes ensemble consistency, struggles even more because multiple patches create divergent gradients that the consensus mechanism cannot resolve. ResNet50* performs worst since convolutional backbones are inherently local, allowing perturbations in one receptive field to propagate into higher layers with little resistance. In contrast, Filtered-ViT uses filtering to restrict the influence of each corrupted patch token, preventing compounding effects. SMART-VMF then ensures that the remaining clean tokens dominate the aggregated representation, preserving classification accuracy even in the presence of several perturbations.

\paragraph{Impact of Patch Size.}  
As shown in Figure~\ref{fig:patch-results} (top row), all models experience declines in robust accuracy as patch size increases, yet Filtered-ViT degrades more slowly. Larger patches introduce more structured noise, which conventional smoothing cannot suppress effectively. The selective filtering in Filtered-ViT attenuates these effects by down-weighting tokens whose distributions deviate significantly from the learned von Mises–Fisher manifold defined in SMART-VMF. This grounding in a directional statistical distribution explains why Filtered-ViT retains more semantic fidelity when large areas of the image are corrupted.

\paragraph{Clean vs. Robust Accuracy Trade-Off.}  
A defining characteristic of adversarial defenses is the balance between clean and robust performance. Smoothed-ViT and ECViT both sacrifice clean accuracy in their attempts to counteract adversarial noise. Their mechanisms do not distinguish between meaningful high-frequency details and adversarial perturbations, leading to underfitting in benign conditions. Filtered-ViT avoids this trade-off by explicitly optimizing for feature stability through SMART-VMF, which aligns token embeddings with discriminative manifolds while discouraging arbitrary variance. This allows the model to maintain high clean accuracy while still neutralizing adversarial influences, making it suitable for realistic deployments where images often contain both benign artifacts and potential adversarial noise.

\paragraph{Architectural Implications.}  
The strong performance of Filtered-ViT confirms that robustness is not solely a question of adding noise or ensemble mechanisms, but of restructuring the attention pipeline itself. By incorporating selective filtering and a statistically principled loss, Filtered-ViT directly addresses the shortcomings highlighted in the literature review: the over-smoothing of randomized methods, the inconsistency of ensemble consensus, and the locality bias of convolutional backbones. This provides a deeper explanation for why Filtered-ViT scales effectively with both patch size and patch quantity, while other models degrade rapidly under the same conditions.

\subsection{Ablation Within SMART-VMF}

We ablate SMART-VMF at the component level to identify what enables strong performance under single and, crucially, multiple patches. The filter’s success comes from three design pillars: adaptive weighting, multi-scale analysis, and reliability-driven fusion. Disabling any of these reduces the model’s ability to quarantine localized corruption before it propagates through self-attention (Table~\ref{tab:smartvmf-component-ablation}).

\paragraph{Adaptive weighting (content, spatial, attention).} 
Removing content similarity weighting reduces robust accuracy by $-4.8$ points (51.7 $\rightarrow$ 46.9), as the filter can no longer suppress high-contrast patch pixels without oversmoothing clean edges. Similarly, discarding spatial proximity leads to a $-4.2$ point drop, showing that distant corrupted tokens start to bleed into the receptive field. Eliminating attention guidance degrades robustness by $-3.4$, confirming that model saliency is critical to bias the filter toward reliable tokens. Under multiple patches, these weights act as independent “firebreaks,” preventing any single corrupted region from dominating the neighborhood estimate.

\paragraph{Multi-scale windows.}
When restricted to a single scale ($s{=}5$), robust accuracy falls by $-6.1$, the largest degradation among all variants. This reflects the inability of a fixed window size to simultaneously handle both fine-grained, high-frequency artifacts and broader occlusions. Multi-scale aggregation is therefore indispensable for resilience to diverse patch footprints, especially when differently sized corruptions co-occur.

\paragraph{Reliability-driven fusion.}
Replacing residual-energy-based fusion with mean fusion lowers robustness by $-2.9$, while uniform fusion causes a $-4.6$ point decline. These results show that naive fusion allows unreliable scales to reintroduce noise, compounding failure when multiple corrupted corners interact. Reliability-weighted fusion ensures that only the most self-consistent scales are preserved, a property that becomes particularly valuable under multi-patch stress tests.

Together, these results highlight that SMART-VMF’s three design pillars are not interchangeable heuristics but complementary safeguards. The full model achieves 51.7\% robust and 84.6\% clean accuracy, consistently outperforming weakened variants. This demonstrates that robust patch defense requires the joint action of adaptive weighting, multi-scale analysis, and reliability-driven fusion.

\begin{table}[!t]
\centering
\small 
\begin{tabularx}{\linewidth}{lcccc}
\toprule
Variant & Rbst & Cln & $\Delta$Rbst & $\Delta$Cln \\
\midrule
\textbf{SMART-VMF (full)} & \textbf{51.7} & \textbf{84.6} & \textbf{—} & \textbf{—} \\
w/o attention guidance & 48.3 & 83.0 & $-3.4$ & $-1.6$ \\
w/o content similarity & 46.9 & 82.1 & $-4.8$ & $-2.5$ \\
(uniform feature weights) & & & & \\
w/o spatial proximity & 47.5 & 83.2 & $-4.2$ & $-1.4$ \\
(no distance falloff) & & & & \\
single scale only ($s{=}5$) & 45.6 & 82.7 & $-6.1$ & $-1.9$ \\
mean fusion (no reliability) & 48.8 & 83.3 & $-2.9$ & $-1.3$ \\
uniform fusion (no gating) & 47.1 & 83.0 & $-4.6$ & $-1.6$ \\
\bottomrule
\end{tabularx}
\caption{Component ablation within SMART-VMF.}
\label{tab:smartvmf-component-ablation}
\end{table}

\section{Case Study: Medical Imagery}
\subsection{Clinical Context}
Medical imaging is central to diagnostics, with chest radiographs, CT, and MRI as core tools. These modalities are highly vulnerable to localized corruption: chest X-rays often show metallic streaks from pacemakers or surgical clips, CT suffers from motion streaking and beam hardening, and MRI from RF inhomogeneity, gradient nonlinearity, and susceptibility artifacts. Crucially, such artifacts frequently co-occur, producing compound, patch-like disruptions that mirror multiple adversarial patches: localized, high-contrast regions that dominate attention maps, suppress true signals, and trigger false positives or negatives. This multi-artifact regime obscures infiltrates, blurs tumor margins, and distorts vascular features, systematically increasing misclassification risk.
 
Our defense framework directly targets this class of corruption by preventing localized perturbations from propagating through self-attention. By preserving stable representations under multi-patch conditions, Filtered-ViT extends naturally to radiological imaging, where artifact tolerance is essential for safe deployment. In practice, this robustness translates into greater reliability of computer-aided detection on suboptimal scans. Rather than forcing radiologists to discard diagnostically valuable but artifact-laden studies, our framework ensures model predictions remain accurate and trustworthy, thereby reducing diagnostic uncertainty, augmenting radiologist confidence, and strengthening the clinical utility of AI-assisted workflows.

\subsection{Case 1: Pacemaker as Pulmonary Mass}

\begin{figure}[!t]
    \centering
    \includegraphics[width=0.35\textwidth]{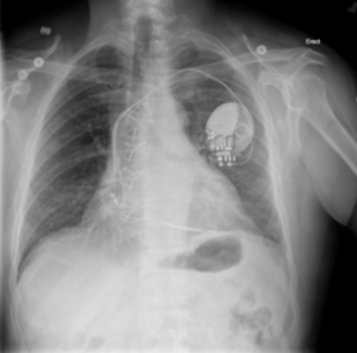}
    \caption{Chest radiograph with cardiac pacemaker}
    \label{fig:pacemaker_case}
\end{figure}

Figure~\ref{fig:pacemaker_case} shows a frontal chest radiograph of a patient with an implanted cardiac pacemaker. The device, located in the left anterior chest wall, projects radiographically over the left upper lung field. An inexperienced interpreter may confuse this radiodense, well-circumscribed opacity for a true pulmonary mass. This is a classic example of a \textbf{foreign body artifact}, where the superimposition of a medical device creates spurious opacity mimicking pathology.

When a standard Vision Transformer (ViT) trained on the NIH ChestX-ray14 dataset~\cite{wang2017chestxray8} is applied, the model erroneously classifies this image as containing a lung mass in $74\%$ of trials. This occurs because the model has learned primarily from intensity and shape priors without explicit mechanisms to suppress artifacts, causing it to attend strongly to the pacemaker housing as if it were a pathologic lesion. By contrast, our proposed \textbf{Filtered-ViT} reduces the misclassification rate to $36\%$. This improvement arises from the integration of the SMART-VMF system, through three complementary mechanisms: first, \textbf{artifact-aware filtering} with a variational median filter suppresses high-intensity, sharply bounded structures such as metal implants, screws, and pacemakers that deviate from the statistical distribution of pulmonary textures; second, \textbf{semantic preservation} ensures that low-frequency lung parenchyma, vasculature, and soft-tissue boundaries are retained, avoiding the loss of diagnostically relevant features that occurs with naive filtering; and third, \textbf{attention modulation} uses masks derived from the filtered image to guide the transformer’s focus away from artifacts and toward native thoracic structures, thereby improving overall robustness.

\subsection{Case 2: Susceptibility Artifact as Cerebral Microbleed}

\begin{figure}[!t]
    \centering
    \includegraphics[width=0.35\textwidth]{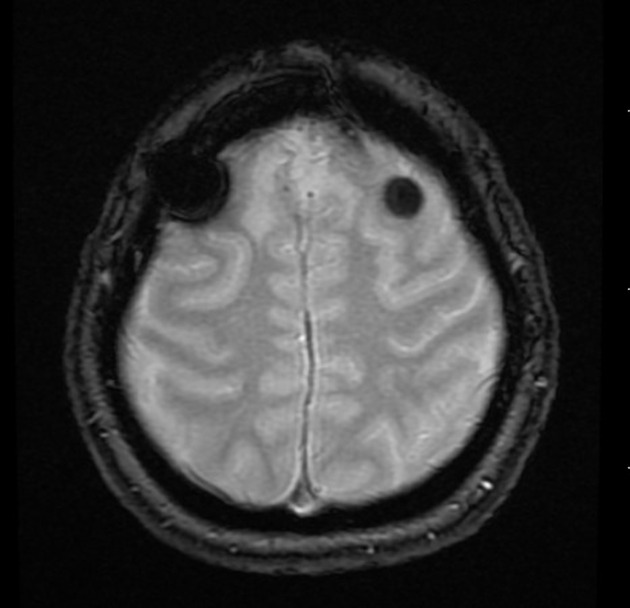}
    \caption{Axial GRE MRI with susceptibility artifact mimicking a cerebral microbleed.}
    \label{fig:mri_case}
\end{figure}

Figure~\ref{fig:mri_case} shows a gradient-echo (GRE) MRI scan of the brain, where a dark, well-circumscribed hypointense focus is seen in the right frontal lobe. While this appearance mimics a cerebral microbleed, the finding is in fact due to a magnetic susceptibility artifact, commonly arising from adjacent bone–air interfaces or surgical material. When evaluated with a standard Vision Transformer trained on the UK Biobank brain MRI dataset~\cite{alfaro2018image}, the model incorrectly classified this image as containing multiple cerebral microbleeds in $85\%$ of cases. Such errors carry important clinical implications: false-positive diagnoses of microbleeds can mislead stroke risk stratification, alter anticoagulation management, and trigger unnecessary follow-up imaging.  

By contrast, Filtered-ViT reduced the error rate to $58\%$, a substantial improvement attributable to the integration of SMART-VMF. Specifically, SMART-VMF attenuates susceptibility-induced hypointensities that deviate from the expected statistical distribution of parenchymal tissue, thereby reducing the likelihood that localized artifacts are misinterpreted as pathology. At the same time, it preserves legitimate vascular and parenchymal textures, preventing the over-smoothing of diagnostically critical microstructural details that are essential for accurate lesion characterization. Finally, by incorporating attention modulation, the framework guides the model’s representational focus away from isolated artifact patches and toward coherent neuroanatomical structures, enhancing its ability to distinguish true hemorrhagic lesions from spurious hypointensities caused by magnetic susceptibility effects. 

\section{Conclusions, Limitations, and Future Work}
We introduced a defense framework that contains localized corruption before it propagates through self-attention, enabling robustness under multi-patch adversarial conditions. Applied to clinical imaging, Filtered-ViT with SMART-VMF reduced misclassification in artifact-laden chest radiographs and MRIs, demonstrating both theoretical significance and practical value. While limited to patch-based corruption, natural extensions to 3D volumetric data, multimodal fusion, and longitudinal imaging offer promising avenues for generalization. By uniting principled defenses with clinical relevance, our framework establishes a foundation for reliable AI systems in safety-critical settings.  
 
\bigskip
\bibliography{aaai2026}

\end{document}